\documentclass[11pt]{article}

\usepackage{amsmath,amsfonts,bm}

\def\eqref#1{equation~\ref{#1}}

\def\1{\bm{1}}

\def\rvh{{\mathbf{h}}}

\def\rvx{{\mathbf{x}}}

\def\vmu{{\bm{\mu}}}

\def\vb{{\bm{b}}}

\def\vx{{\bm{x}}}
\def\vy{{\bm{y}}}

\def\mW{{\bm{W}}}

\DeclareMathAlphabet{\mathsfit}{\encodingdefault}{\sfdefault}{m}{sl}
\SetMathAlphabet{\mathsfit}{bold}{\encodingdefault}{\sfdefault}{bx}{n}

\usepackage{PRIMEarxiv}

\usepackage{amsthm}
\usepackage{times}
\usepackage{epsfig}
\usepackage{amsmath}
\usepackage{algorithm}
\usepackage{algorithmicx}  
\usepackage{algpseudocode} 
\usepackage{booktabs}       
\usepackage{multirow}
\usepackage{graphicx}
\usepackage{wrapfig}
\usepackage{multirow}

\usepackage[utf8]{inputenc} 
\usepackage[T1]{fontenc}    
\usepackage{hyperref}       
\usepackage{url}            
\usepackage{booktabs}       
\usepackage{amsfonts}       
\usepackage{nicefrac}       
\usepackage{microtype}      
\usepackage{xcolor}         

\usepackage{caption}
\usepackage{subcaption}
\usepackage[round]{natbib}

\newtheorem{lemma}{Lemma}

\begin{document}

\title{Rethinking Normalization Methods in Federated Learning}








\author{
  Zhixu Du\thanks{Equal contribution}\\
   Duke University \\
  \texttt{zhixu.du@duke.edu} \\
   \And
  Jingwei Sun\footnotemark[1] \\
  Duke University \\
  \texttt{jingwei.sun@duke.edu} \\
  \And
  Ang Li \\
  Duke University \\
  \texttt{ang.li630@duke.edu} \\
  \And
  Pin-Yu Chen \\
  IBM Research AI \\
  \texttt{pin-yu.chen@ibm.com} \\
  \And
  Jianyi Zhang \\
  Duke University \\
  \texttt{jianyi.zhang@duke.edu} \\
  \And
  Hai "Helen" Li \\
  Duke University \\
  \texttt{hai.li@duke.edu} \\
  \And
  Yiran Chen \\
  Duke University \\
  \texttt{yiran.chen@duke.edu} \\
  }

\maketitle

\begin{abstract}
Federated learning (FL) is a popular distributed learning framework that can reduce privacy risks by not explicitly sharing private data. In this work, we explicitly uncover \textbf{external covariate shift} problem in FL, which is caused by the independent local training processes on different devices. We demonstrate that external covariate shifts will lead to the obliteration of some devices' contributions to the global model. Further, we show that normalization layers are indispensable in FL since their inherited properties can alleviate the problem of obliterating some devices' contributions. However, recent works have shown that batch normalization, which is one of the standard components in many deep neural networks, will incur accuracy drop of the global model in FL. The essential reason for the failure of batch normalization in FL is poorly studied. We unveil that external covariate shift is the key reason why batch normalization is ineffective in FL. We also show that layer normalization is a better choice in FL which can mitigate the external covariate shift and improve the performance of the global model. We conduct experiments on CIFAR10 under non-IID settings. The results demonstrate that models with layer normalization converge fastest and achieve the best or comparable accuracy for three different model architectures.

\end{abstract}

\section{Introduction}

Federated  learning  (FL)~\cite{mcmahan2017communication,tang2021fedgp}  is  a  popular  distributed learning approach that enables a large number of devices to train a shared model in a federated fashion without explicitly sharing  their local data. 
In order to reduce communication cost, most FL methods enable participating devices to conduct multiple steps of training before uploading their local models to the central server for aggregation. However, multiple steps of local training on edge devices would cause \textit{internal covariate shift}~\cite{ioffe2015batch} on local models, which is a known problem in the centralized (non-FL) setting.
Internal covariate shift describes the phenomenon that during the training of deep neural networks (DNN), each layer's input distribution varies due to the parameter changes of preceding layers. 
Such an issue requires the internal neurons in a given layer to adapt to varying input distributions, and hence slows down the convergence of model training. 


\begin{figure}[ht]
\centering
     \includegraphics[scale=0.45]{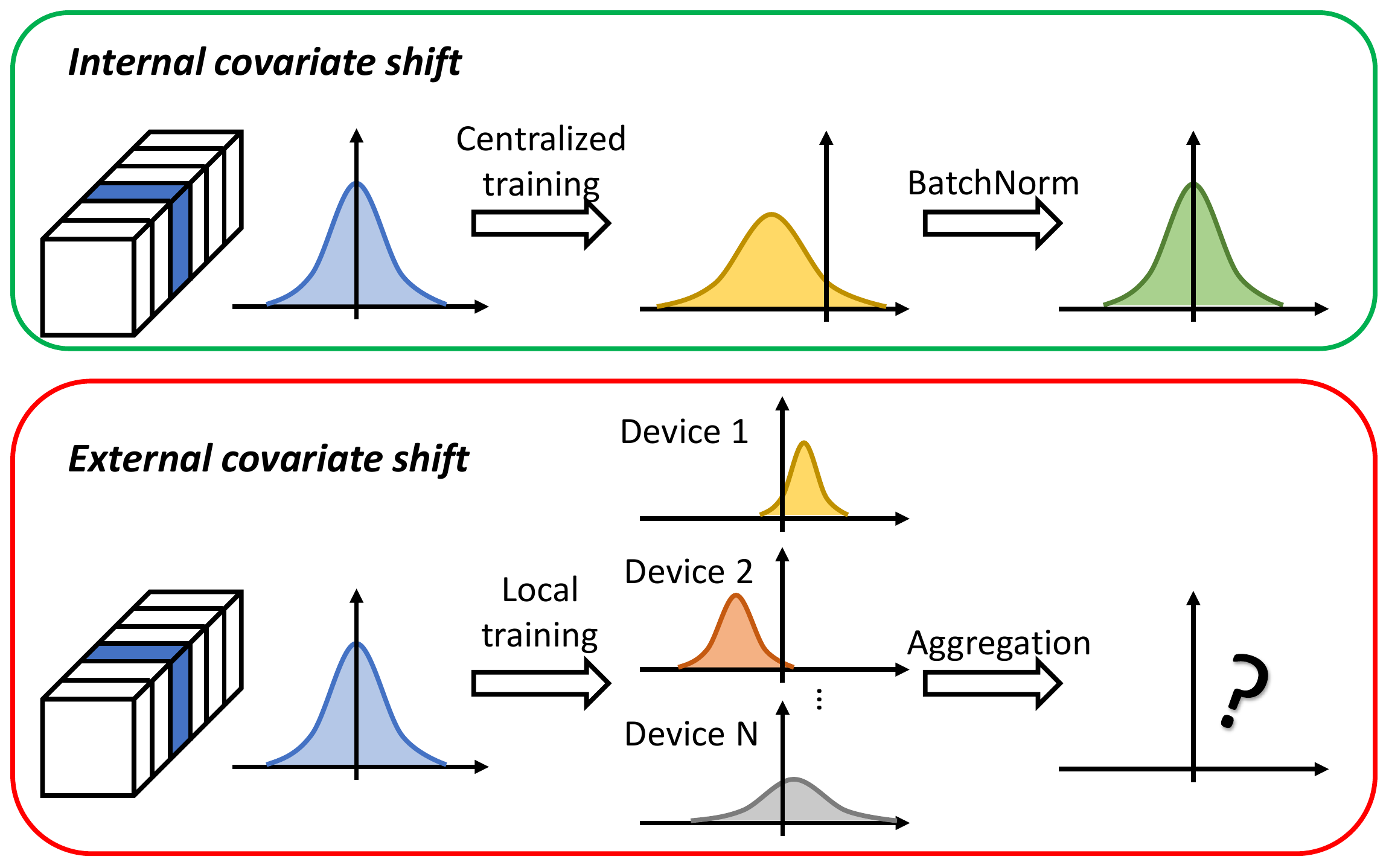}
\caption{Internal and external covariate shift.}
\label{fig:covariate_shift}
\end{figure}

 Internal covariate shift has been well studied in the centralized learning scenarios and an effective approach to mitigate this issue is batch normalization. Further, batch normalization has many good properties which will stable the training process exploited by previous work. In FL systems, participating devices perform several batches of local training in each communication round, thus, internal covariate shift raises a concern for the local training. In FL, the updates of model parameters vary across devices during local training. Without any constrains, the internal covariate shift across devices will be varied, leading to gaps of statistics information given the same channel among different devices. We name this unique phenomenon in FL as \textit{external covariate shift}. Due to external covariate shift, the model neurons of a given channel on one device need to adapt to the feature distribution of the same channel on other devices, 
which slows down the convergence of global model training. Further, external covariate shift may also lead to large discrepancy in the norm of weights and may obliterate contribution from devices with weights of small norm. 
 
 We show in this paper that inherited good properties of normalization will shed light on solving external covariate shift. However, existing works~\cite{li2021fedbn,hsieh2020non} show that batch normalization will incur the accuracy drop of global model in FL. These works simply attribute the failure of batch normalization in FL to the discrepancies of local data distributions across devices. In this work, we show our key observation that the ineffectiveness of batch normalization in FL is not only caused by the data distribution discrepancies, but also resulted from the diverged internal covariate shift among different devices due to the stochastic training process. Batch normalization drops the accuracy of global model when applied to solve external covariate shift because the feature distribution of the global model after aggregation is not predictable. Further, we also show that layer normalization does not suffer from the problem and can server as the placement of batch normalization in FL.

  The experiment results demonstrate that layer normalization can effectively mitigate the external covariate shift and speedup the convergence of the global model training. In particular, layer normalization achieves the fastest convergence and best or comparable accuracy upon convergence on three different model architectures.

Our key contributions are summarized as follows:

\begin{itemize}
    \item To the best of our knowledge, this is the first work to explicitly reveal \textit{external covariate shift} in FL, which is an important issue that affects the convergence of FL training.
    \item We propose a simple yet effective placement of batch normalization in Federated Learning, i.e., layer normalization, which can effectively mitigate the external covariate shift and speedup the convergence of FL training.
\end{itemize}
\section{Preliminaries}
\label{sec:prelim}
\subsection{Internal covariate shift and Activation normalization}
In the training of deep neural networks, each layer's input distribution keeps changing due to updates of parameters in the preceding layer. Consequently, layers are forced to keep adapting to the varying input distributions, leading to slow convergence. The issue is more severe when networks get deeper, because the covariate shift will be amplified layer by layer. This phenomenon is called \textit{internal covariate shift}~\cite{ioffe2015batch}. Activation normalization is proposed to alleviate the internal covirate shift. 

Activation normalization methods have become one of the most important components in Deep Neural Networks (DNNs) aiming at alleviating internal covariate shift. The normalization layer is usually inserted between neural networks' output layer and activation functions. Among activation normalization methods, batch normalization~\citep{ioffe2015batch}, group normalization~\citep{wu2018group} and layer normalization~\citep{ba2016layer} are the most commonly used. Activation normalization methods are of great help in stabilizing the training of DNNs and producing well-conditioned training landscapes~\citep{martens2015optimizing}. Specifically, activation normalization methods usually have two running statistics $\vmu$ and $\bm{\sigma}$ and two trainable parameters $\gamma$ and $\beta$ for scaling and shifting, following the same formula,
\begin{equation}
    \textnormal{AN}(\rvx) = \frac{\rvx-\vmu}{\sqrt{\bm{\sigma}^2+\epsilon}}\times\gamma + \beta,
\end{equation}
where AN stands for Activation Normalization and $\rvx$ is the input of the normalization. Different normalization techniques computes $\vmu$ and $\bm{\sigma}$ differently. Batch normalization uses mini-batch mean and mini-batch variance as $\vmu$ and $\bm{\sigma}$, while layer normalization and group normalization use mean and variance across channels or features of a data sample. 

Previous works~\citep{ioffe2015batch, ba2016layer} have exploited many well-inherited properties of batch normalization and layer normalization, and some of them are important in Federated Learning, which will be detailed in the following sections.

\subsection{Scale invariant property of normalization}
\label{sec:scale_inv}
The scale invariant property of batch normalization (BN) and layer normalization (LN) has been widely studied~\citep{neyshabur2016, sun2020new, van2017l2, arora2018theoretical}. The property equips BN and LN layers with ability to automatically tune the learning rate for the layer that are antecedent to the normalization layer. Formally, 
\begin{equation}
    \textnormal{BN}(\rvh; \mW) = \textnormal{BN}(\rvh; a\mW),
\end{equation}
where $\mW$ is the weight parameters of the preceding layer, $\rvh$ is the input to the layer and $a$ is a non-zero scalar. The same also applies for LN. In back-propagation, 
\begin{equation}
    \label{eq:auto_tuning}
    \frac{\partial \textnormal{BN}(\rvh; a\mW)}{\partial a\mW} = \frac{1}{a}\times\frac{\partial \textnormal{BN}(\rvh; \mW)}{\partial \mW}.
\end{equation}
If the weight $a\mW$ is large, then in back-propagation the gradient will be shrink by a factor of $a$. On the other hand, if  $a\mW$ is small, the gradient will be enlarged. 

Further, the scale invariant property leads to an equilibrium on the norm of weights. The following lemma shows that the gradient on batch normalization and layer normalization layer is always orthogonal to the weight parameters. 
\begin{lemma}
    If function $f$ satisfies that $f(\lambda\mW)=f(\mW)$ for all non-zero scalar $\lambda$ and $\nabla f(\mW)$ exists, then $$\mW^\top \nabla f(\mW) = 0,$$ where $\nabla f(\mW)$ is the gradient of $f$ with regard to $\mW$. 
\end{lemma}
Lemma 1 can be easily proved by differentiating with regard to $\lambda$ on both sides of the equation and set $\lambda=1$. By applying Lemma 1 and the rule of weights update
\begin{equation}
    \mW_{t+1} = \mW_{t} + \eta\frac{\partial f(\mW_{t})}{\partial \mW_{t}},
\end{equation}
we can derive that
\begin{equation}
    \label{eq:norm_weight}
    \lVert\mW_{t+1}\rVert_2 = \lVert\mW_{t}\rVert_2 + \eta^2\lVert\frac{\partial f(\mW_{t})}{\partial \mW_{t}}\rVert_2,
\end{equation}
where the middle term disappears because of the property in Lemma 1. The auto-tuning effect in Eq. \ref{eq:auto_tuning} will force the norm of weights to converge to an equilibrium. When the norm of weights is large, the norm of gradients will be small and vice versa. We will discuss why this property is important in Federated Learning in the next section. 

\section{External Covariate Shift and Adaptive-Balancing of Weights}

\subsection{External convariate shift}
In Federated Learning, participating devices train their local models for multiple steps in each communication round with their own private data. In this process, we observe that the statistics of channels is significantly diverse between devices. This phenomenon is caused by the independent local training process on different devices,  resulting in different internal covariate shifts. We name this phenomenon as \textit{external covariate shift}, which is a unique problem in FL.

The external covariate shift phenomenon slows down the convergence of the global model in Federated Learning, since the feature distribution varies after aggregation and neurons have to keep adapting the updated feature distribution. We attribute this feature distribution shifts as the key reason that batch normalization drops the global accuracy, which will be detailed in Section~\ref{sec:bn}. 

\begin{figure}[h!]
\centering
     \includegraphics[scale=0.45]{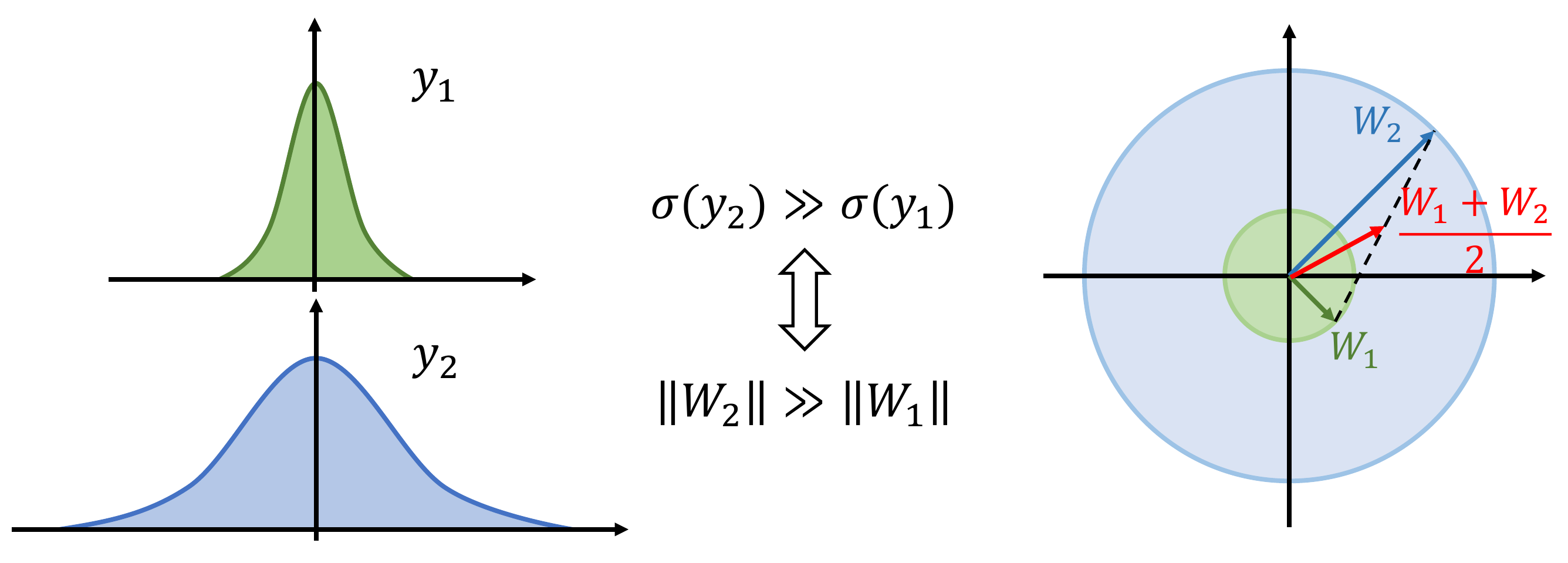}
\caption{The contribution of the model with a smaller norm will be obliterated by the model with a larger norm during aggregation.}
\label{fig:rethink}
\end{figure}

In addition to the obstacle to convergence caused by the heterogeneous feature distributions, the external covariate shift phenomenon also harms the aggregation step in the federated training process. To describe why the aggregation step is harmed by the external covariate shift, we consider a toy example. For simplicity, we use the fully connected (FC) layer as an example for analysis. Note that such an analysis can be naturally extended to other types of layers. Specifically, a FC layer is represented as:
\begin{equation}
    \vy = \mW \vx + \vb,
\end{equation}
where $\vx$ is the input to this layer, $\mW$ and $\vb$ are weight and bias, $\vy$ is the feature of this layer. We assume that the input features $\vx$ for this layer are whitened (independently distributed with zero mean and unit variance), then $\mu[\vy]=\mu[\vb]$ and $\sigma[\vy]=||\mW||_2$ where $||\cdot||_2$ denotes the Euclidean norm. For two sets of weights $\mW_1$ and $\mW_2$, if their feature deviations have significant discrepancies, i.e., $\sigma[\vy_1] \gg \sigma[\vy_2]$, then we can derive that $||\mW_1||_2\gg||\mW_2||_2$. Thus, the essence of external covariate shift describes the shift of model parameter's norm. Considering these two sets of weights belonging to two local models involved in FL, then the contribution of $\mW_2$ will be obliterated by $\mW_1$ as shown in Figure~\ref{fig:rethink}, and the same effect applies to the bias, which is more related to $\mu[\vy]$.

In FL, it is reasonable that weights on different devices are diverse due to non-IID data, but they should have similar norms $||\mW||_2$, otherwise the aggregated model would be dominated by a small part of local models with larger weight norms.

\subsection{Adaptive-balancing on norm of weights}
The scale invariant property inherited in the batch normalization and layer normalization is able to alleviate the obliteration problem in aggregation caused by external covariate shift. In Section \ref{sec:scale_inv}, we demonstrate that the batch and layer normalization can automatically tune the learning rate of the weights of the previous layer. Consequently, the norm of weights will reach certain equilibrium and this process is regardless of the input distribution. Since in federated learning, we start training with the same initialization, the scale invariant property provides some guarantee that the norm of weights across different devices will not diverge far away from initialization. Formally, we iteratively apply Eq.~\ref{eq:norm_weight}, we can see that 

\begin{equation}
    \lVert\mW_{t+1}\rVert_2 = \lVert\mW_{0}\rVert_2 + \eta^2\sum_{i=0}^t\lVert\frac{\partial f(\mW_{i})}{\partial \mW_{i}}\rVert_2,
\end{equation}
where $\mW_{0}$ is the initialization. Combined with Eq.~\ref{eq:auto_tuning}, we have the norm of gradient will be similar since otherwise the auto tuning property will automatically magnify or reduce the norm of weights. The adaptive-balancing property makes the normalization module indispensable in Federated Learning. However, the batch normalization has some issues in the distributed training scenario.

\subsection{Failure of batch normalization}
\label{sec:bn}
Batch normalization has been the most commonly used normalization layer in deep neural networks that satisfies the scale invariant property. However, recent studies~\cite{li2021fedbn, hsieh2020non} have shown that applying batch normalization to FL incurs accuracy drop. Though previous arts have raised the problem, a convincing explanation is lacking, where the previous work simply blame the data heterogeneity. Although different data samples will generate totally different features, it is still reasonable that the whole channel of outputs has similar statistics. This is the foundational assumption of batch normalization, because it indeed normalizes the output of different samples by applying the same statistics in a given channel. In particular, for two devices holding data sampled from different distributions, the neurons in their models may follow totally different distributions, but the statistics of the given channel is not necessarily different between two devices. In this case, batch normalization can still be applied to effectively address the internal covariate shift in FL.

Therefore, we explain the phenomena by lens of external covariate shift. Due to external covariate shift, model neurons have to adapt to the new input distributions after aggregation. This is caused by the different input distributions of the corresponding neurons on other devices, which we call \textit{external neurons}. After applying batch normalization, different devices will have varying running statistics, and the central server can not obtain correct running statistics by simply averaging local statistics. Wrongly obtained batch normalization statistics that mismatch feature statistics will lead to information loss or introduces extra noise to the features~\cite{gao2021representative} especially after activation functions. Therefore, based on our observation, we identify the key reason that why batch normalization causes accuracy drop in FL is that the statistics of the same channel are trained to be different between devices during local training.

\begin{figure*}[h!]

\centering
    \resizebox{0.95\linewidth}{!}{
     \includegraphics{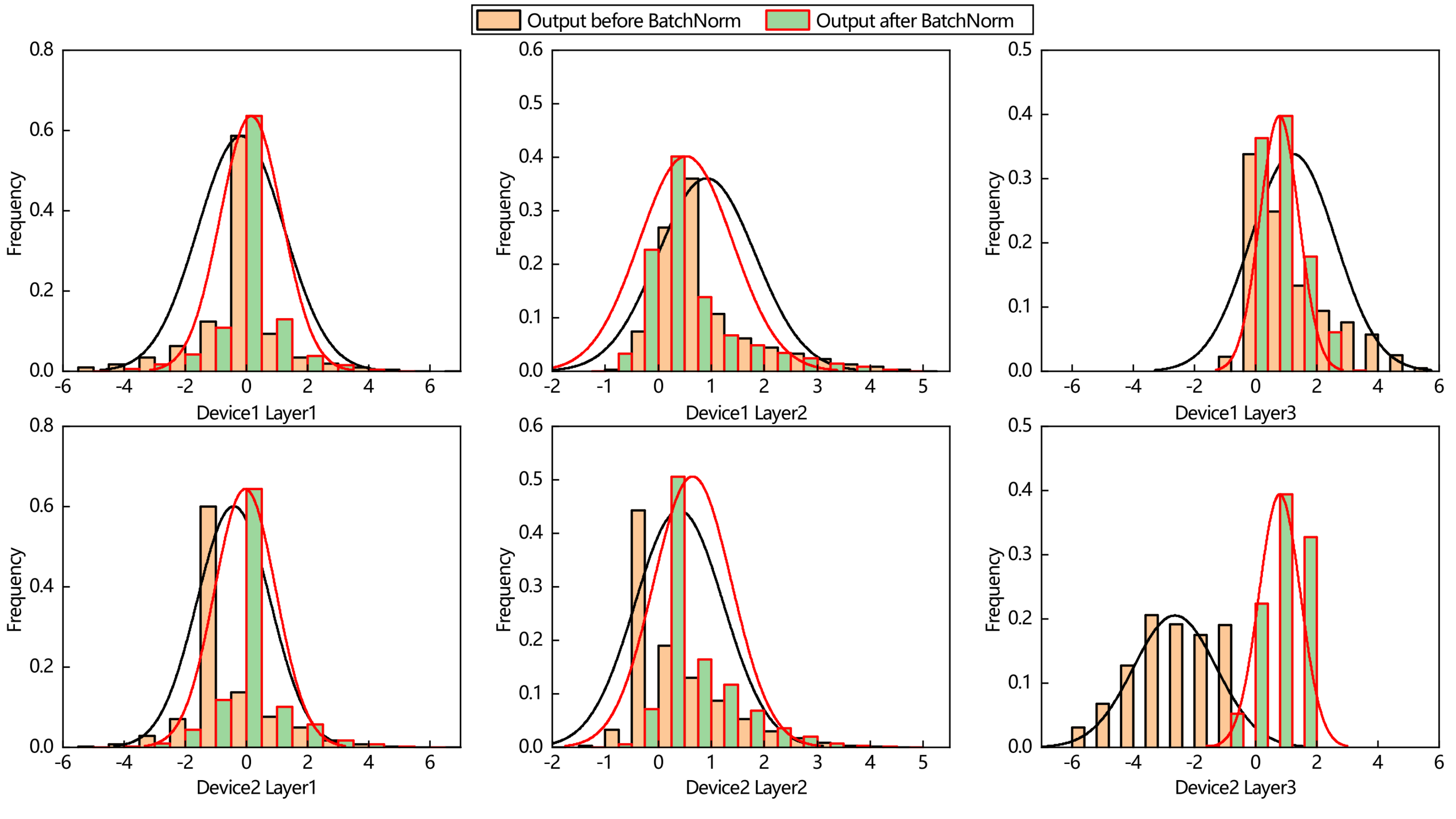}
     }
\caption{Histograms of output before and after BatchNorm in different channels on two devices.}
\label{fig:external_covariate_shift}
\end{figure*}

We verify our observation through a toy experiment. The histograms of output from different channels on two devices are shown in Figure~\ref{fig:external_covariate_shift} on experiments on MNIST. In particular, we train two identical models from the same initialization with two local training dataset to simulate two devices in a federated learning round. For simplicity, we apply a base model with 3 convolutional layers followed by batch normalization layers respectively, and for each layer there is only one channel. To avoid the influence of non-IID data, the two local training dataset are the same except having different mean values, which can be easily normalized by batch normalization. 

Note that the output after batch normalization (the red curve) is the input of the succeeding layer. The results show that two devices have totally different feature statistics for the same channel (layers) with training from the same initialized model. 
For the second and third layer, the input distributions are normalized to similar statistics (the red curves in layer1 and layer2 figures), but the output channels (the black curves in layer2 and layer3 figures) still show a significant statistical discrepancy. 

\begin{table}[ht]
\centering
\caption{CNN Model architecture.}
\begin{tabular}{c}
  \toprule
  5$\times$ 5 Conv 3-6 \\
  \hline
  NormLayer \\
  \hline
  5$\times$ 5 Conv 6-16 \\
  \hline
  NormLayer \\
  \hline
  FC-120 \\
  \hline
  NormLayer\\
  \hline
  FC-84 \\
  \hline
  FC-10 \\
  \bottomrule
\end{tabular}
\label{table:model}
\end{table}

\subsection{Success of layer normalization}
Given the failure of batch normalization, layer normalization becomes the good choice of the normalization layer in Federated Learning. Since layer normalization normalizes across channel in a sample-wise manner, it does not suffer from the statistics discrepancy across devices. Further, layer normalization possesses all the advantages discussed in Section~\ref{sec:prelim}. In next section, we verify our analysis through experiments and we show that layer normalization is the best among all baselines.

\section{Experiments}

In our experiments, we utilize FedAVG~\citep{mcmahan2017communication} as the algorithm and apply different normalization methods on different architectures, i.e., VGG-11, ResNet-18 and a simple CNN. Each experiment is run with three different random seeds and standard deviation is reported. Experiments are conducted on a server with two Intel Xeon E5-2687W CPUs and four Nvidia TITAN RTX GPUs.

\subsection{Experimental Setup}
\paragraph{Dataset.}
We use CIFAR10 dataset and construct non-IID dataset by following the configurations in~\cite{mcmahan2017communication}. The data is distributed across 100 devices. Each device holds 2 random classes with 100 samples per class. 
\paragraph{Hyperparameter configurations.}
We set local epoch $E$ as 10 and batch size $B$ as 64. We apply SGD optimizer and set the learning rate $\eta$ to 0.01. In each communication round, there are 10 randomly sampled  devices participate in the training. The architecture of CNN model is presented in Table\ref{table:model}. By default, we perform training with 5000 communication rounds.
\paragraph{Baselines}
We compare the performance of different normalization methods, group normalization, layer normalization, instance normalization and batch normalization. Further, we run experiments on fixed batch normalization where the training statistics are not aggregated in the aggregation step. We also provide results with no normalization as baselines.

\subsection{Experimental Results}

\begin{figure*}
    \centering
    
      \begin{subfigure}[b]{0.33\textwidth}
             \centering
             \includegraphics[width=\textwidth]{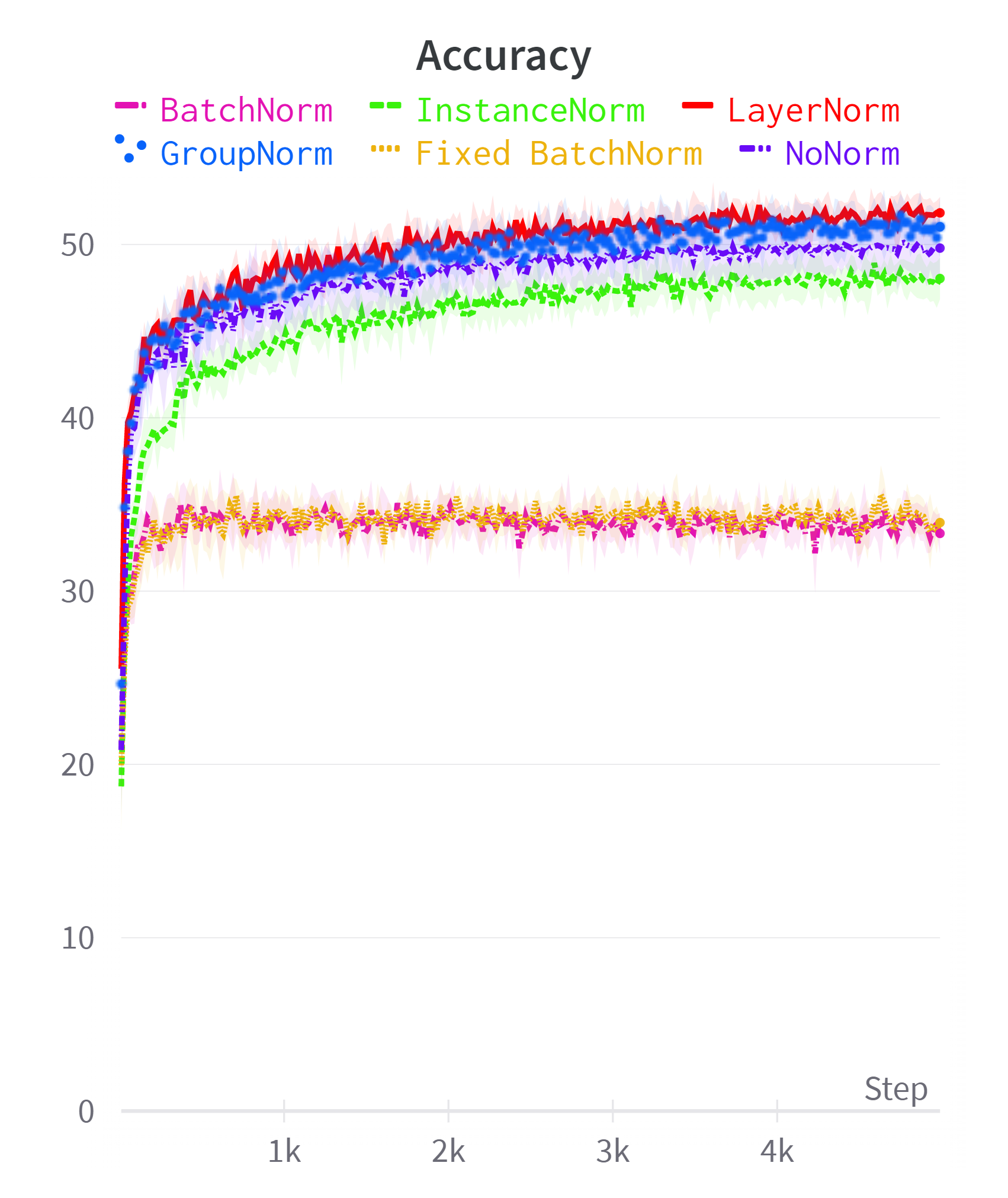}
             \caption{CNN}
             \label{fig:CNN}
     \end{subfigure}
     \begin{subfigure}[b]{0.33\textwidth}
             \centering
             \includegraphics[width=\textwidth]{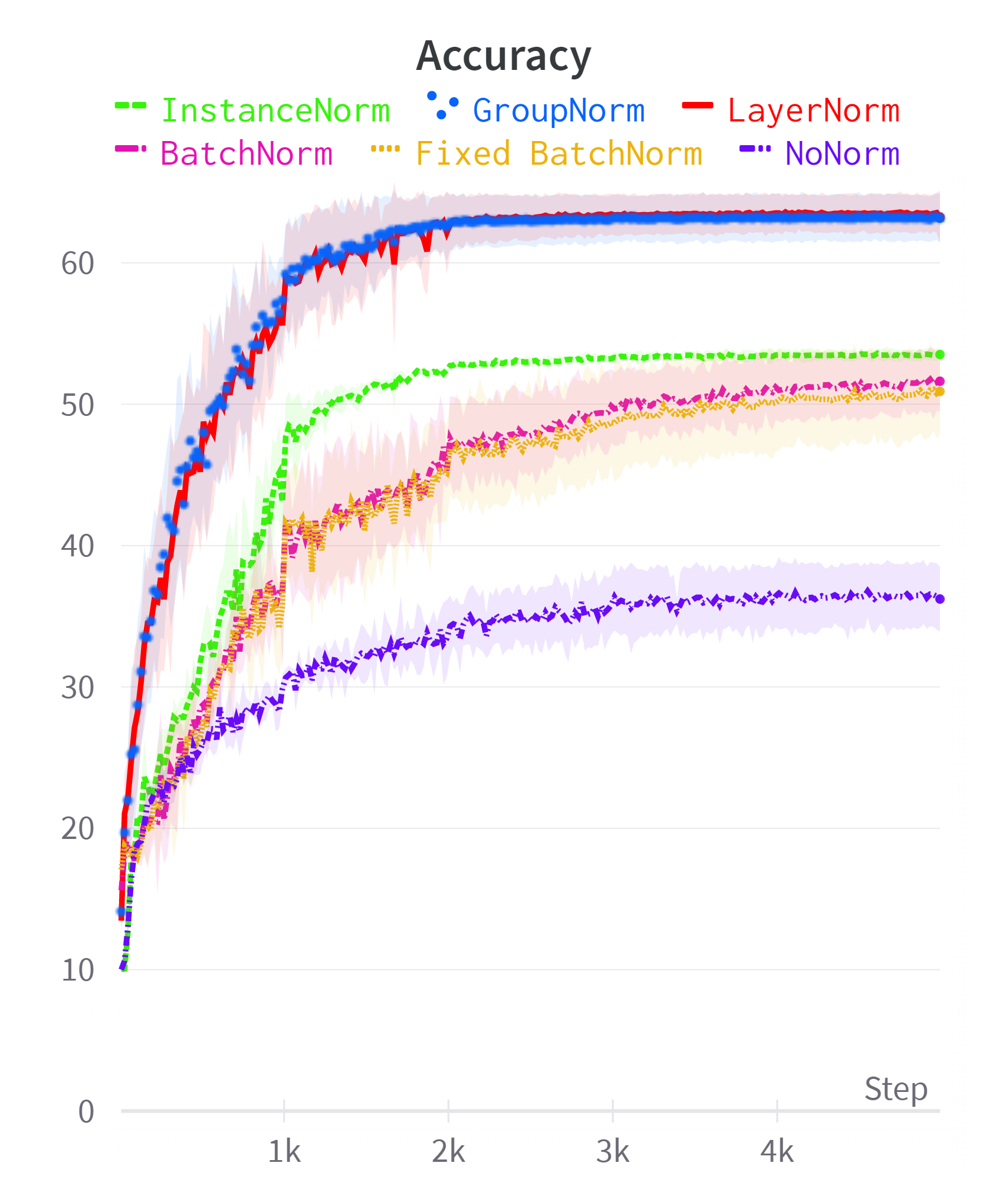}
             \caption{VGG-11}
             \label{fig:VGG}
     \end{subfigure}
     \begin{subfigure}[b]{0.33\textwidth}
             \centering
             \includegraphics[width=\textwidth]{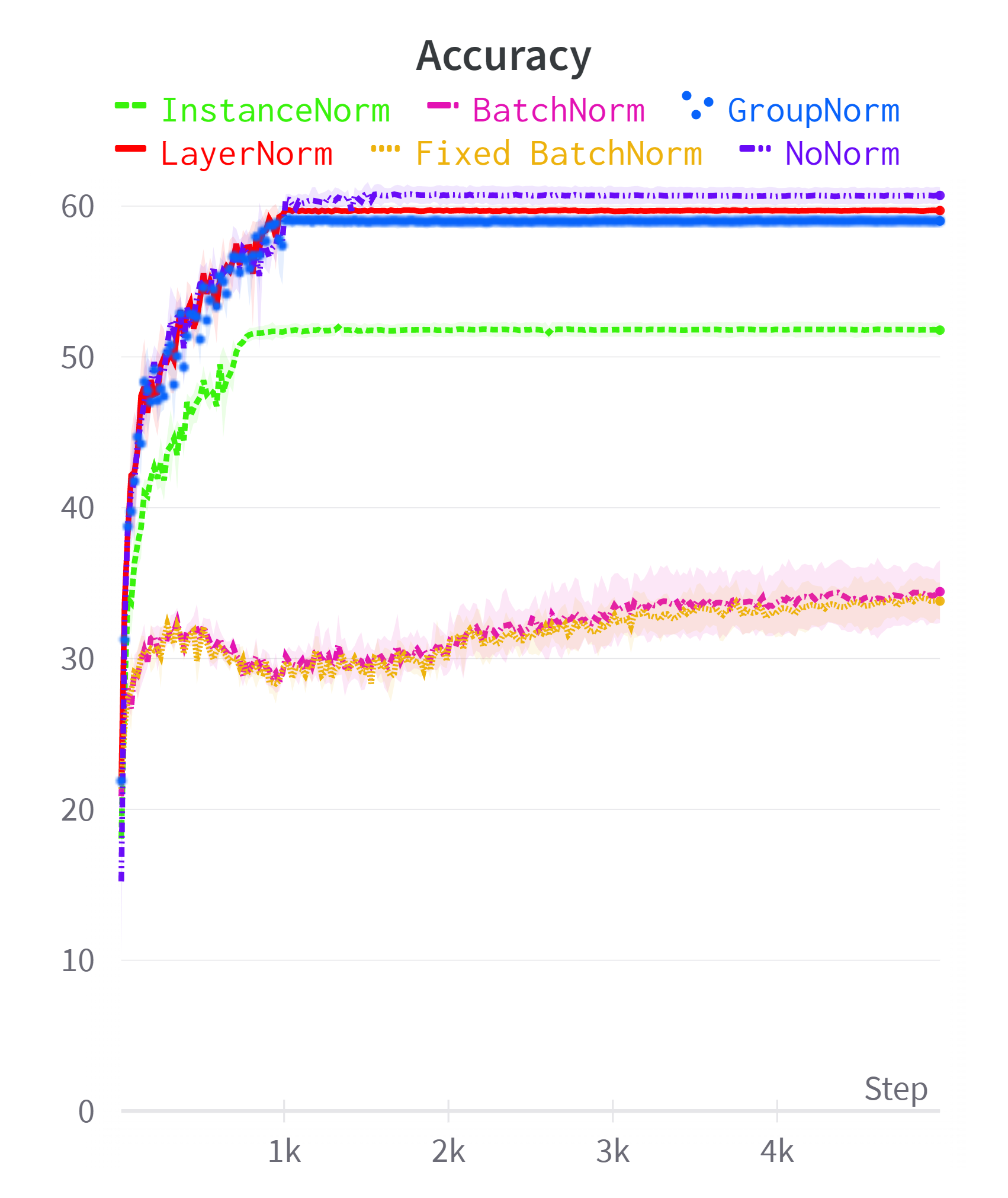}
             \caption{ResNet-18}
             \label{fig:ResNet}
     \end{subfigure}
     \caption{We show the convergence of different normalization method. Layer normalization achieves fastest or comparable convergence and best or comparable accuracy. }
    \label{fig:performance}
\end{figure*}

The comparison of convergence speed is shown in Figure~\ref{fig:performance} and the accuracy upon convergence is shown in Table~\ref{tab:acc}. The results show that layer normalization converges the fastest and to the highest accuracy or comparable to the highest accuracy for all three architectures. Note that, in all experiments, group normalization achieves similar results as layer normalization since these two methods are very similar. For VGG-11, the layer normalization achieves 11.96\% and 27.01\% improvements compared with batch normalization and no normalization, respectively. For ResNet and CNN, layer normalization also achieves remarkable improvements compared with batch normalization. Note that, for ResNet-18, the no normalization achieves the best result which is slightly higher than layer normalization, we account this for the residual connection. Since the residual connection, the input to the next block is not too far away from the input to the last block, which prevents the weights deviating from the initialization. However, the hyper-parameters for ResNet with no normalization must be selected carefully to obtain a stable training process.

\makeatletter
\newcommand\figcaption{\def\@captype{figure}\caption}
\newcommand\tabcaption{\def\@captype{table}\caption}
\makeatother

\begin{table}[t]
\caption{Accuracy on CIFAR10 with different normalization methods on different architectures. Standard deviation is computed with regard to three different  seeds.}
\centering
\label{tab:acc}
{
\begin{tabular}{clc}
\toprule[1pt]
\centering
\multirow{2}{*}{Architecture}&
\multirow{2}{*}{Method}&
\multicolumn{1}{c}{Accuracy($\%$)~($\uparrow$)}\\

&&{@5000 rounds} \\
\toprule[1pt]

\multirow{6}{*}{VGG-11} &No normalization& $36.51_{\pm 2.48}$\\

&Group-Normalization & $63.10_{\pm 1.68}$\\

&Layer-Normalization& \bfseries $63.52_{\pm 1.44}$\\

&Instance-Normalization& $53.46_{\pm 0.34}$\\

&Batch-Normalization& $51.56_{\pm 2.22}$\\

&Fixed Batch-Normalization & $50.93_{\pm 3.07}$\\

\toprule[1pt]

\multirow{6}{*}{ResNet-18}&No normalization& \bfseries $60.73_{\pm 0.53}$\\

&Group-Normalization &$59.01_{\pm 0.44}$\\

&Layer-Normalization& $59.70_{\pm 0.21}$\\

&Instance-Normalization& $51.78_{\pm 0.47}$\\

&Batch-Normalization& $34.38_{\pm 2.07}$\\

&Fixed Batch-Normalization &$33.85_{\pm 1.25}$\\

\toprule[1pt]

\multirow{6}{*}{CNN}&No normalization& $49.96_{\pm 1.80}$\\

&Group-Normalization &$50.86_{\pm 1.08}$\\

&Layer-Normalization& \bfseries $52.02_{\pm 0.76}$\\

&Instance-Normalization& $48.18_{\pm 1.17}$\\

&Batch-Normalization& $33.29_{\pm 0.66}$\\

&Fixed Batch-Normalization & $33.66_{\pm 1.34}$\\

\toprule[1pt]

\end{tabular}
}
\vspace{1em}
\end{table}

\section{Related Work}

\paragraph{Batch Normalization in Deep Neural Networks}

Batch Normalization (BatchNorm) was proposed to solve the internal covariate shift problem in training deep neural networks. It has been shown that BatchNorm can effectively speedup and improve the robustness of model training. Relevant works have explained why BatchNorm improve both convergence and generalization in training neural networks. \cite{luo2018towards} investigated an explicit regularization form of BN and illustrates the advantage of applying BatchNorm in a single layer perceptron. \cite{santurkar2018does} demonstrates that BatchNorm makes the optimization landscape significantly smoother and this smoothness induces a more predictive and stable behavior of the gradients, allowing for faster training. \cite{morcos2018importance} empirically shows that BatchNorm  improves generalization by implicitly discouraging single direction reliance of the model. In addition to improving generalization and convergence, BatchNorm is also applied~\cite{li2018adaptive} to tackle the domain adaptation problem. However, recent works~\cite{li2021fedbn,hsieh2020non} show that BatchNorm incurs the accuracy drop of the global model in FL.

\paragraph{Normalization in Federated Learning}

BatchNorm is shown to be ineffective in FL, and recent works propose several alternatives. FedBN~\cite{li2021fedbn} applies local batch normalization to alleviate the feature shift before averaging models by not uploading and averaging local batch normalization parameters during central aggregation. However, FedBN is limited to personalized FL scenarios. \cite{hsieh2020non} demonstrates that group normalization (GroupNorm) can improve the convergence of FL. Nevertheless, GroupNorm is instance-based normalization, which is highly sensitive to the noise on data samples. HeteroFL\cite{diao2020heterofl} applied statistic BatchNorm to solve the privacy concern by not tracking running estimates and simply normalize batch data. 
Although these alternatives empirically shows better performance than BatchNorm in FL, the essential reasons why BatchNorm is ineffective in FL are still poorly studied.
\section{Conclusion}
In this paper, we explicitly identify the external covariate shift problem in FL, which is caused by not only non-IID data but also independent training processes on different devices. We also demonstrate that severe external covariate shift even  obliterates  some devices' contributions to the global model, which will significantly  degrade FL training performance. We further present the importance of scale invariant property of normalization layer to the Federated Learning, i.e., prevent norm of weights on different devices from deviating the initialization. We empirically and theoretically explain that external covariate shift is the key reason why batch normalization incurs accuracy drop of the global model in FL and we show that layer normalization does not suffer the problem. The experimental results demonstrate that layer normalization converges much faster than other normalization methods and achieve the best or comparable to the best accuracy.

\bibliographystyle{ACM-Reference-Format}
\bibliography{reference}

\end{document}